\documentclass[12pt,a4paper]{article}
\usepackage{authblk}
\usepackage[utf8]{inputenc}
\usepackage{graphicx} 

\title{Exploring Autonomous Agents through the Lens of Large Language Models: A Review}
\author{Saikat Barua}
\affil{North South University} 

\date{February 2024} 

\begin{document}

\maketitle

\begin{abstract}
Large Language Models (LLMs) are transforming artificial intelligence, enabling autonomous agents to perform diverse tasks across various domains. These agents, proficient in human-like text comprehension and generation, have the potential to revolutionize sectors from customer service to healthcare. However, they face challenges such as multimodality, human value alignment, hallucinations, and evaluation. Techniques like prompting, reasoning, tool utilization, and in-context learning are being explored to enhance their capabilities. Evaluation platforms like AgentBench, WebArena, and ToolLLM provide robust methods for assessing these agents in complex scenarios. These advancements are leading to the development of more resilient and capable autonomous agents, anticipated to become integral in our digital lives, assisting in tasks from email responses to disease diagnosis. The future of AI, with LLMs at the forefront, is promising.
\end{abstract}

\newpage
\tableofcontents

\newpage
\section{Introduction}

From antiquity, the quest for automation has been a constant in human endeavors, driven by the desire for increased productivity and efficiency. The genesis of autonomous agents is rooted in early conceptualizations of self-governing systems capable of intentional action in the physical world\cite{franklin1996agent}. This paradigm has found applications across a spectrum of fields, from cognitive science to economics. The evolution of autonomous agents has seen significant strides, with the advent of large language models (LLMs) marking a pivotal moment in our pursuit of artificial intelligence that mirrors human cognition\cite{wang2023survey}. The ripple effects of automation on society are far-reaching. It has catalyzed the emergence of novel work methodologies and has reshaped societal constructs. However, it’s crucial to acknowledge the existence of an “automation gap”. The reality is that not all tasks are amenable to automation, and in such instances, human intervention remains indispensable to fuel the engine of innovation\cite{chui2015four}. The dawn of automation has revolutionized the work landscape. While it has amplified productivity, it has also marginalized less-educated workers and augmented the monopoly rents accrued by capital owners. Interestingly, even occupations at the top of the economic pyramid, such as financial managers, physicians, and senior executives, encompass a significant proportion of activities that are susceptible to automation\cite{manyika2018ai}. While automation has been a catalyst for economic growth, it has concurrently widened the chasm of wealth inequality. The fruits of automation are not equitably distributed, and the wealth generated often gravitates toward the upper echelons of society\cite{koru2020automation}. This has ignited discourse on the societal ramifications of automation and the imperative for policy interventions to ensure equitable wealth distribution. The amalgamation of large language models with autonomous agents offers a promising frontier for enhancing simulation capabilities\cite{gao2023large}. LLMs, endowed with a wealth of web knowledge, have shown extraordinary promise in approximating human-level intelligence \cite{wang2023survey}. This has spurred a surge in research exploring the potential of LLM-based autonomous agents\cite{boiko2023emergent}.

LLMs, being trained on extensive internet data, encapsulate a substantial corpus of human knowledge. This mirrors the Semantic Web’s objective of rendering internet data machine-readable\cite{berners17may}, thereby forging a web of data amenable to machine processing. The interaction dynamics between humans and LLMs are crucial. The manner in which we query or prompt these models can significantly shape the responses. This brings into focus the strategy of prompt tuning, a technique employed to enhance LLM performance by carefully selecting and adjusting prompts or seed texts to guide the model’s generated text\cite{reynolds2021prompt}\cite{peng2023model}. The learning process of LLMs, driven by interaction with data, offers a pathway to deciphering human cognition\cite{shaki2023cognitive}\cite{binz2023turning}. Researchers straddling the domains of artificial intelligence and cognitive neuroscience are exploring whether these computational models can serve as proxies for language processing in the human brain. The emergence of LLMs has provided a window into the world of general-purpose autonomous agents\cite{cheng2024exploring}. These agents, powered by LLMs, demonstrate robust generalization capabilities across a range of applications, functioning as autonomous general-purpose task assistants.The integration of LLMs and multimodal models not only augments the agent’s capabilities but also bestows upon it the semblance of a silicon lifeform\cite{wang2024large}. For embodied tasks, where robots interact with complex environments, text-only LLMs often encounter challenges due to a lack of compatibility with robotic visual perception. However, the fusion of LLMs and multimodal models into various robotic tasks offers a holistic solution\cite{wang2024large}\cite{mandi2023roco}.

Unleashing the Potential of LLMs in Autonomous Agents This review paper aims to delve into the integration of Large Language Models (LLMs) into autonomous agents, a transformative process that has given birth to intelligent agents capable of executing tasks that were deemed unattainable for decades. For instance, DeepMind’s AlphaGo, through self-play and learning from its own errors, honed its proficiency in the game of Go\cite{hassabis2016alphago}. Similarly, AlphaFold, another brainchild of DeepMind, showcased its prowess in predicting protein folds with astounding accuracy, thereby resolving a longstanding grand challenge in biology\cite{team2020alphafold}. LLM-based Agents and Their Effective Functioning The purview of this review encompasses agents built on the foundation of LLMs using techniques like LangChain\cite{Chase_LangChain_2022} and AutoGPT\cite{Significant_Gravitas_AutoGPT}, which necessitate precise protocols for effective functioning. While these techniques have demonstrated potential in constructing automated research assistants, they have also encountered hurdles such as prompt tuning issues in AutoGPT and the need for more efficient token consumption in LangChain. These challenges present opportunities for improvement and refinement. The review also grapples with the limitations and challenges of employing different types of LLMs in agent construction, as well as the possibilities they present. For instance, LLaMA\cite{touvron2023llama}, an open-source LLM, has found application in agent construction. Conversely, closed-source models have also been deployed for similar purposes. The promise of open-source LLMs is noteworthy, as they democratize tool access, foster transparency, and stimulate innovation. Drawing an analogy to operating systems, the open-source design of Linux is often deemed more efficient than its counterparts, Windows and Mac, in terms of performance. This suggests that open-source LLMs, leveraging their design and community support, could eventually outperform their closed-source counterparts. The fusion of LLMs with autonomous agents has ushered in a new era in the realm of AI. Despite the challenges that persist, the continuous advancements and the growing inclination towards open-source models hint at a promising future for this technology.

\section{Background on Large Language Models and LLM-based Autonomous Agents}

\subsection{Transformer Architecture}
Three distinct types of transformer architectures underpin LLMs and Their Underlying Transformer Architecture Large Language Models (LLMs):

\textbf{Encoder-Decoder:} The pioneering transformer architecture, where the encoder processes the input sequence to generate a hidden representation, which the decoder then uses to produce the desired output sequence. T5\cite{DBLP:journals/corr/abs-1910-10683} and BART\cite{DBLP:journals/corr/abs-1910-13461} are notable models that employ this architecture.

\textbf{Encoder-only:} This architecture is employed when the task requires only the encoding of the input sequence, eliminating the need for a decoder. The input sequence is encoded into a fixed-length representation, which is then used as input to a classifier or a regressor for prediction. BERT\cite{devlin2019bert}, DistilBERT\cite{sanh2020distilbert}, and RoBERTa\cite{liu2019roberta} are models that utilize this architecture.

\textbf{Decoder-only:} This architecture comprises solely a decoder, which is trained to predict the subsequent token in a sequence given the preceding tokens. The key distinction between the Decoder-only and the Encoder-Decoder architectures is that the former lacks an explicit encoder to summarize the input information.

\begin{figure}[htp]
    \centering
    \includegraphics[scale=0.5]{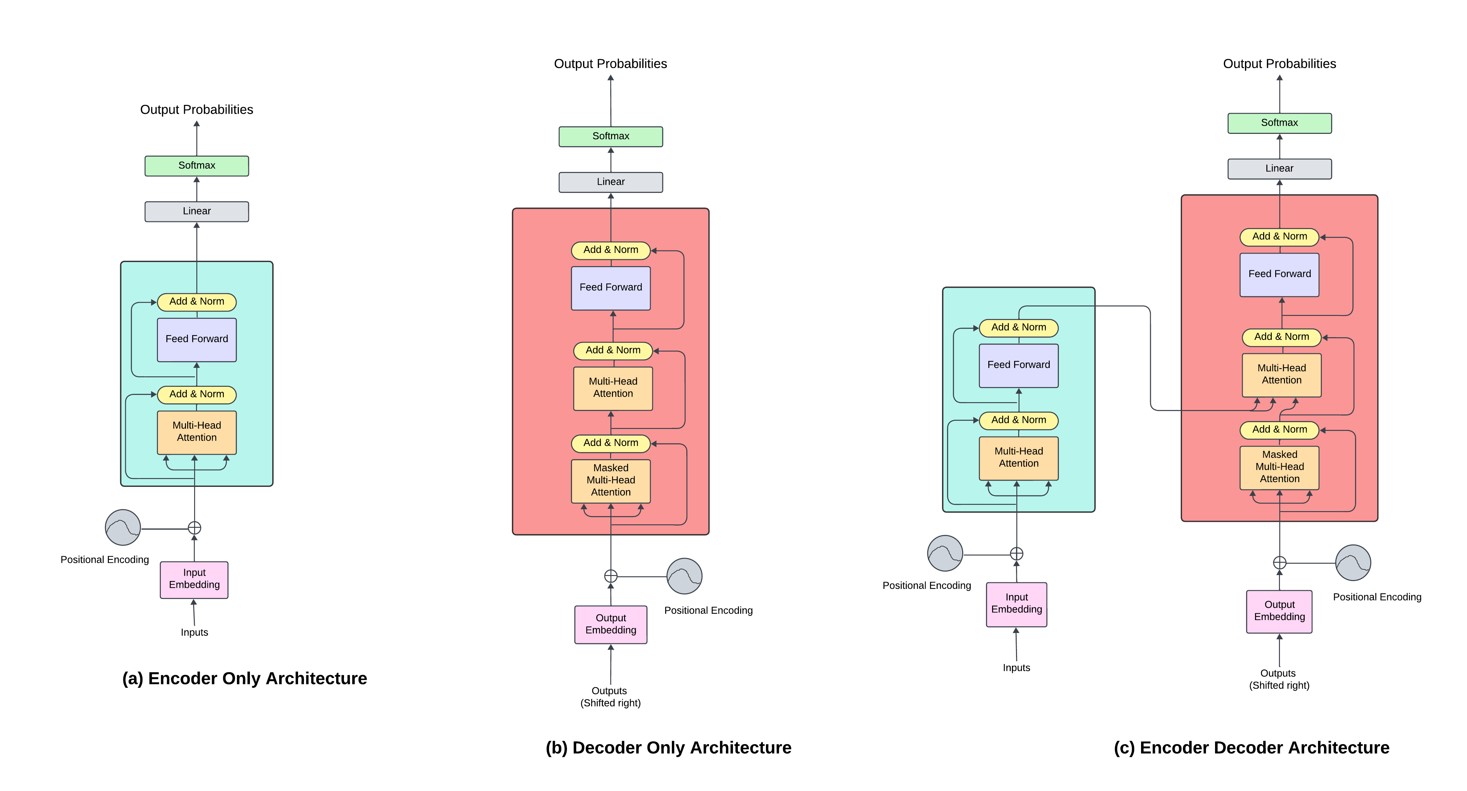}
    \caption{Architecture of Transformer (Based on \cite{vaswani2023attention})}
    \label{fig:my_label}
\end{figure}

Figure 1 presents the Architecture of the Transformer model. Transformers have outperformed Recurrent Neural Networks (RNNs) in numerous aspects. RNNs process data sequentially\cite{Sherstinsky_2020}, resulting in slower processing times and limited capabilities, particularly with longer data sequences. In contrast, transformers process sentences in their entirety rather than word by word, enabling parallel computation and reducing training time. Furthermore, transformers are built on the concept of self-attention, allowing the model to assign importance to different words in a sentence based on their relevance to each other.

\subsection{The Evolution of Large Language Models}

The Evolution and Development of LLMs The genesis of LLMs can be traced back to early research in natural language processing and machine learning. However, their rapid evolution was sparked by the advent of deep learning techniques and the introduction of the Transformer architecture. The development of LLMs using the transformer architecture significantly enhanced the performance of natural language tasks compared to the previous generation of RNNs, leading to a surge in regenerative capability.

Developed by OpenAI, GPT-4\cite{achiam2023gpt} is among the most advanced LLMs currently available. It has demonstrated remarkable capabilities, including complex reasoning understanding, advanced coding capability, and proficiency in multiple academic exams. LLaMa 2\cite{touvron2023llama} is a groundbreaking addition to their AI model lineup. It’s designed to power a range of state-of-the-art applications. LLaMa 2’s training data is vast and varied, marking a significant advancement over its predecessor. T5\cite{DBLP:journals/corr/abs-1910-10683}, developed by Google, is an encoder-decoder model designed for flexibility and can be fine-tuned for a variety of tasks. BART\cite{DBLP:journals/corr/abs-1910-13461}, developed by Facebook, is a denoising sequence-to-sequence pre-training model designed for natural language generation, translation, and comprehension.
RoBERTa\cite{liu2019roberta}, a variant of BERT that employs a different pre-training approach, has been shown to outperform BERT on several benchmarks.
Each of these models has its own strengths and limitations. For instance, while GPT-4 has demonstrated impressive capabilities, it’s slow to respond, and the inference time is much higher, prompting developers to resort to the older GPT-3.5 model. Similarly, while LLaMa 2 boasts vast and varied training data, its availability was restricted to prevent misuse. Despite their prowess, T5, BART, and RoBERTa are not without their limitations. For instance, T5 has been observed to falter on out-of-distribution testing examples when compared to in-distribution testing examples. BART, meanwhile, grapples with constraints in processing long sequences, with a maximum sequence length of 512. RoBERTa, despite its substantial performance enhancements, presents challenges in conducting meticulous comparisons between different methodologies due to computational costs, varying sizes of private datasets, and the significant influence of hyperparameter selections.

The confluence of autonomous agents and Large Language Models (LLMs) is a rapidly evolving domain, where each LLM can be viewed as an agent within an ecosystem, collaborating to accomplish a wide array of tasks. The architecture of these LLM-based autonomous agents can be dissected into Memory, enabling the agent to remember past actions and situate itself within a dynamic environment; Planning, where the agent decomposes high-level instructions or goals into manageable tasks, orchestrates their execution, assesses the outcomes, and refines them to fulfill the instruction or goal as accurately as possible; and Action, where the agent exhibits the capacity to comprehend software call syntax and semantics, selects the necessary software tools for any task, and operates them by providing syntactically and semantically correct parameters. Several successful implementations of LLM-based autonomous agents have been documented, including Chemcrow\cite{bran2023chemcrow}, which uses LLMs for chemistry-related tasks; MathAgent\cite{liao2023modeling}, which employs LLMs to solve intricate mathematical problems; Autogen\cite{wu2023autogen}, which utilizes LLMs to autonomously generate content; Libro\cite{kang2023large}, a literature-focused LLM-based agent; ChatLaw\cite{cui2023chatlaw}, which provides legal advice using LLMs; and Proagent\cite{zhang2024proagent}, which uses LLMs to carry out professional tasks.

The integration of Large Vision Models (LVMs) in Autonomous Agents is a rapidly advancing research domain. Autonomous Vision Agents, equipped with a profound understanding of images, can utilize the redundant parameters of LVMs to discern the scope of an image and apply overparameterization when their inherent feature detection falls short. Hu et al.\cite{hu2023avis} unveils AVIS, an autonomous framework for visual question answering that seeks information. AVIS harnesses a Large Language Model (LLM) to dynamically strategize the use of external tools and scrutinize their outputs, thereby acquiring the essential knowledge needed to respond to the questions posed. Wang et al.\cite{wang2023review} offers a comprehensive survey of the methodologies used in the computer vision domain for large vision models and visual prompt engineering. It delves into the latest breakthroughs in visual prompt engineering and showcases influential large models in the visual domain. Zhou et al.\cite{zhou2023vision} explores the applications of Vision-Language Models (VLMs) in Autonomous Driving (AD) and Intelligent Transportation Systems (ITS). It underscores the exceptional performance of these models and their capacity to leverage Large Language Models (LLMs).

Large Speech Models (LSMs) are a specialized category of Large Language Models (LLMs) that are trained on extensive speech data. Yu Zhang et al. present the Universal Speech Model (USM)\cite{zhang2023google}, a unified model capable of performing automatic speech recognition (ASR) in over 100 languages. The USM is developed by pre-training the model’s encoder on a substantial unlabeled multilingual dataset, comprising 12 million hours of data across more than 300 languages, and subsequently fine-tuning it on a smaller labeled dataset. The authors employ multilingual pre-training, random-projection quantization, and speech-text modality matching to achieve leading performance in multilingual ASR and speech-to-text translation tasks. Notably, the USM demonstrates comparable or superior performance to the Whisper model in both in-domain and out-of-domain speech recognition tasks across numerous languages, despite utilizing a labeled training set that is one-seventh the size. The Autonomous agents infused LSM have been a focal point of research in both academia and industry, leading to a surge in studies exploring LLM-based autonomous agents. Large Language and Speech Model (LLaSM)\cite{shu2023llasm} can follow speech-and-language instructions, offering a more user-friendly and natural way for humans to interact with artificial intelligence. Fathullah 
et al.\cite{fathullah2023prompting}investigate the augmentation of large language models (LLMs) with speech recognition capabilities. They achieve this by integrating a compact audio encoder with the LLM, transforming it into an automatic speech recognition (ASR) system. Their experiments on the Multilingual LibriSpeech (MLS) dataset demonstrate that the integration of a conformer encoder with the open-source LLaMA-7B model enables it to outperform monolingual baselines by 18\% and conduct multilingual speech recognition, despite the model’s predominant training on English text. These studies underscore the potential of LSMs in augmenting the capabilities of autonomous agents, particularly in the areas of noise filtering and robust training processes.

The fusion of Large Language Models (LLMs) with autonomous agents is a rapidly evolving research area, particularly in the sphere of explainability. Zhao et al.\cite{zhao2023expel} present a paradigm shift from traditional learning models by introducing Experiential Learning (ExpeL) agents that learn from their experiences without necessitating parametric updates. Concurrently, the importance of explainability in AI and autonomous agents is emphasized by Anjomshoae et al.\cite{anjomshoae2019explainable}, advocating for goal-driven explainable AI to foster trust and understanding in increasingly complex AI systems. Crouse et al.\cite{crouse2023formally} underscore the need for novel methodologies that enable learning from agent experiences, enhancing their performance and effectiveness. Furthermore, Barua et al.\cite{barua2023kaxai} unveils a system that amalgamates AutoML, XAI, and synthetic data generation, offering users a seamless experience while navigating the power of machine learning. The paper also introduces LLM-based synthetic data generation for improving model performance, reliability, and interoperability.

\section{Building Autonomous Agents with Large Language Models}

\subsection{All About Memory}

Large Language Models (LLMs) employ a diverse memory architecture, primarily utilized to store the model’s parameters and intermediate computational activations. Notably, in transformer-based LLMs, the key-value cache memory for each request can be substantial and dynamically fluctuate in size. To efficiently manage this, some systems adopt techniques inspired by classical virtual memory and paging methodologies prevalent in operating systems.

When designing memory systems, engineers frequently draw inspiration from nature. Neural networks, a prominent example of biological systems inspiring computer algorithms, were initially mathematically conceptualized as a system of simplistic neurons capable of executing simple logical operations\cite{bongard2009biologically}. Cognitive psychology provides foundational frameworks, such as Baddeley’s multi-component working memory model, which have been pivotal in understanding human memory\cite{guo2023empowering}. Certain LLMs employ memory management techniques inspired by operating systems. For instance, PagedAttention\cite{kwon2023efficient}, an attention algorithm inspired by the classical virtual memory and paging techniques in operating systems, has been proposed to manage the key-value cache memory in LLMs.

The memory complexity in LLMs is a consequence of their design. In transformer-based LLMs, the memory requirement escalates quadratically with the input sequence length due to the self-attention mechanism in transformers computing pairwise interactions between all tokens in the input.

In the realm of LLM memory management, memories are typically organized in hierarchies. For example, in the MemGPT model\cite{packer2023memgpt}, context windows are treated as a constrained memory resource, and a memory hierarchy for LLMs is designed analogous to memory tiers used in traditional operating systems. This hierarchy aims to provide greater continuity for nuanced contextual reasoning during intricate tasks and collaborative scenarios. On a typical machine, the memory hierarchy comprises three levels. The higher levels are faster but scarce, while the lower levels are slower but abundant.

Large Language Models (LLMs) employ a diverse memory architecture, primarily utilized to store the model’s parameters and intermediate computational activations. Notably, in transformer-based LLMs, the key-value cache memory for each request can be substantial and dynamically fluctuate in size. The fundamental building blocks of all LLMs closely follow this pattern, where the layers are integrated with different permutations.

\begin{itemize}

    \item Recurrent Layers: These layers preserve information from preceding inputs, facilitating the processing of sequences and temporal dependencies. Notable examples include Long Short-Term Memory (LSTM) and Gated Recurrent Units (GRU).

    \item Feedforward Layers: These layers execute non-linear transformations on input data, thereby extracting salient features and patterns.

    \item Embedding Layers: These layers transcribe words or tokens into dense vectors, encapsulating semantic relationships.

    \item Attention Layers: These layers selectively concentrate on pertinent parts of the input, thereby enhancing task performance and contextual comprehension. They allocate weights to different tokens based on their significance, enabling LLMs to capture long-range dependencies and relationships within the text\cite{alizadeh2024llm}.
    
\end{itemize}

Side by Side, Novel Architectures of LLMs are capable of preserving dialogue context for coherent conversations, tailoring responses based on user preferences, and tracking factual information for knowledge-based tasks\cite{modarressi2023retllm}.

Empirical investigations of LLMs often concentrate on their capacity to process and comprehend natural language. They transform free-form text inputs into arrays of numbers, known as embeddings, which are lower-dimensional, numerical representations of the original text that aim to capture the underlying linguistic context. Further Research has demonstrated the proficiency of Large Language Models (LLMs) in acquiring linguistic patterns and representations from extensive text corpora. The employed encoding structures significantly influence performance and the capacity for generalization. The incorporation of multi-head attention and profound transformer architectures has been instrumental in achieving cutting-edge outcomes in numerous Natural Language Processing tasks\cite{huang2023llms}\cite{guo2023gpt4graph}.

Different LLMs may employ distinct encoding strategies. For instance, transformer-based models heavily rely on multi-head attention for encoding. Models such as BERT, GPT-3, T5, and BARD utilize multi-head attention with varying configurations. LLaMA, on the other hand, amalgamates recurrent layers (LSTM) with attention mechanisms. Multi-head attention involves parallel attention layers attending to different parts of the input, capturing diverse relationships. Models like GPT-3, T5, BARD, and LLaMA prominently utilize multi-head attention. Feedforward attention, conversely, involves focused attention on specific parts of the input, filtering out irrelevant information. It is often employed within transformer layers for enhanced feature extraction. Models like GPT-2, BERT, and XLNet use this type of attention. The number of layers in an LLM escalates its model complexity and adaptability to handle intricate language patterns. For example, GPT-3 has 96 layers, T5 has 110 layers, BARD has 66 layers, and LLaMA has 75 layers. There are trade-offs between performance, computational cost, and memory requirements.

Within the sphere of language model generation, the selection of a decoding strategy is instrumental in determining the output. These strategies, which guide how a model chooses the subsequent word in a sequence, can markedly affect the coherence, diversity, and overall caliber of the generated text. Noteworthy among these strategies are:

\begin{itemize}
    \item Greedy Search: This approach opts for the most likely token at each juncture, which can often result in repetitive or conservative responses.

    \item  Multinomial Sampling: This method introduces an element of randomness to foster diversity, although it may yield incoherent text. 
    
    \item Beam Search: This technique maintains a record of multiple potential sequences, striking a balance between quality and diversity. 
    
    \item Contrastive Search: This strategy aims for clarity and consistency, minimizing ambiguity and repetition. 
    
\end{itemize}

GPT-3 typically employs multinomial sampling for the generation of creative text. T5 and BARD frequently utilize beam search for tasks necessitating accuracy and coherence. LLaMA investigates the use of contrastive search to enhance clarity and consistency\cite{ning2023skeletonofthought}.

\subsection{From LLM with Autonomous Agents}

Large Language Models have exhibited extraordinary proficiency in assimilating linguistic patterns and representations from vast text corpora. Nevertheless, their efficacy in executing tasks necessitating planning and action is constrained. For instance, while LLMs can convert objectives articulated in natural language into a structured planning language, they grapple with tasks involving numerical or physical reasoning\cite{xie2023translating}. Tasks of this nature encompass solving ordinary differential equations as observed in celestial mechanics, forecasting the movements of planets, stars, and galaxies, and numerical linear algebra in data analysis. Furthermore, the action sequence in LLMs is frequently hardcoded, restricting their adaptability. For instance, in LangChain, chains represent a hardcoded sequence of actions, whereas agents employ LLMs to determine which action sequence to undertake.

To circumvent the limitations of LLMs in task execution, researchers have investigated the concept of multimodality, which entails the integration of diverse types of data (e.g., text, images, audio) into the model. However, multimodality alone cannot resolve issues such as arithmetic, geometry, chemistry, math equations, etc. These tasks often demand specific reasoning abilities or domain knowledge that cannot be captured by merely processing multiple types of data. This is where the concept of tools becomes pertinent. Tools are the fundamental building blocks that augment the capabilities of LLMs. They are executable units with specific functions, enabling LLMs to perform a variety of tasks. By amalgamating different tools, LLMs can create a flow that achieves a broad spectrum of goals. For instance, an LLM agent has access to tools that can perform actions beyond text generation, such as conducting a web search, executing code, performing a database lookup, or doing math calculations.

LangChain\cite{Chase_LangChain_2022} is a comprehensive library engineered to facilitate the construction of applications powered by LLMs. It enables the creation of complex interaction flows with LLMs by chaining together different components from several modules. In LangChain, an agent is formed by amalgamating tools and memory. The fundamental concept of agents in LangChain is to employ an LLM to select a sequence of actions to undertake. Depending on the user input, the agent can then decide which, if any, of these tools to invoke. In essence, an Agent in LangChain equals Tools plus Memory. Upon receiving a request, Agents utilize an LLM to decide on which action to undertake. After an action is completed, the agent updates its memory, which aids in maintaining the context of the conversation. This approach enables LangChain to manage more complex, structured interactions, maintaining context or sequence between different prompts and responses.

LiteLLM\cite{LiteLLM} is another framework that simplifies interactions with various LLM APIs. It provides a lightweight package in the OpenAI format, streamlining the process of calling API endpoints from providers like OpenAI, Azure, Cohere, and Anthropic. LiteLLM is an excellent choice for anyone looking to swiftly and effortlessly incorporate non-OpenAI models into production.

Langchain and LiteLLM, while both serving as intermediaries that bridge the divide between various Large Language Models (LLMs) and project prerequisites, exhibit differences in their methodologies and functionalities. Langchain adopts more of a builder’s framework, facilitating the creation of intricate interaction flows with LLMs. It offers a more structured approach to chaining prompts and responses, proving particularly beneficial when the maintenance of context or sequence between different prompts and responses is essential. Conversely, LiteLLM streamlines interactions with a variety of LLM APIs, offering a consistent API that enables seamless switching between LLMs without the need for extensive code modifications. Its primary focus lies in providing a lightweight package for invoking API endpoints from various providers.

Auto-GPT\cite{Significant_Gravitas_AutoGPT}, an open-source AI agent, employs OpenAI’s GPT-4 or GPT-3.5 APIs. It strives to achieve a goal expressed in natural language by decomposing it into sub-tasks and utilizing the internet and other tools in an automated loop. As one of the inaugural applications using GPT-4 for autonomous tasks, Auto-GPT demonstrates adaptability and learning capabilities, making it a valuable asset for both businesses and individuals. However, it has been observed that Auto-GPT may occasionally fall into logic loops or "rabbit holes", which can limit its problem-solving abilities. In contrast, while both Auto-GPT and LangChain utilize LLMs, their core functionalities are distinct. Auto-GPT is an autonomous agent that decomposes a large task into various sub-tasks without requiring user input. It can become ensnared in logic loops and “rabbit holes”, thereby limiting its problem-solving capacity in certain scenarios. Conversely, LangChain is not an autonomous agent but rather an LLM library that facilitates the development of a wide range of applications atop state-of-the-art LLMs. It offers simple interfaces and robust backend support for the deployment, scaling, and fine-tuning of these models. This distinction underscores the diverse ways in which LLMs can be harnessed for different purposes.

\subsection{The Art of Reasoning and Acting}

Large Language Models (LLMs) have demonstrated an extraordinary capacity to emulate human-level intelligence, leading to a surge in research exploring LLM-based autonomous agents. Autonomous agents, long considered a promising pathway to achieving Artificial General Intelligence (AGI), are expected to execute tasks through self-guided planning and actions. These agents are typically designed to operate based on simple, heuristic policy functions and are trained in isolated, constrained environments. This approach, however, contrasts with the human learning process, which is inherently complex and capable of learning from a broad range of environments.

LLMs, with their intuitive natural language interface, are ideally suited for user-computer interactions and tackling complex problems. Certain pretrained LLMs, such as GPT-4, possess significant reasoning capabilities, enabling them to deconstruct complex issues into simpler steps, providing solutions, actions, and evaluations at each stage. However, as closed systems, LLMs are unable to access the most recent data or specific domain knowledge, leading to potential errors or “hallucinations” (i.e., generating incorrect responses). Consequently, it becomes crucial to design an autonomous agent in conjunction with LLMs. Direct prompting does not consistently yield thoughtful responses from LLMs. They are prone to hallucinate more if an immediate response is sought\cite{huang2023survey}. Hallucination in LLMs refers to instances where the model generates text that is incorrect, nonsensical, or unreal. As LLMs are not databases or search engines, they do not cite the basis of their responses. These models generate text extrapolated from the provided prompt.

To address this, Single Path Reasoning (SPR) and Multipath Reasoning (MPR) modules have been developed. SPR, exemplified by HuggingGPT\cite{shen2023hugginggpt} and Chain of Thought (CoT)\cite{wei2023chainofthought}, encourages intermediate reasoning steps. CoT generates a series of reasoning steps, marking a significant advancement in this direction. Chain of Thought bolsters the reasoning abilities of Large Language Models (LLMs) by generating a series of intermediate reasoning steps or prompts. It dissects multi-step problems into intermediate stages, enabling additional computation by your models as needed. It proves to be an effective strategy for complex computational problems where conventional methods fall short. HuggingGPT is an LLM-empowered agent that utilizes LLMs to bridge various AI models in machine learning communities to solve AI tasks. It employs ChatGPT to orchestrate task planning upon receiving a user request, selects models based on their function descriptions available in Hugging Face, executes each subtask with the chosen AI model, and summarises the response according to the execution results.

MPR, in contrast, facilitates automatic multiple interactions between users and LLMs by using previously generated answers as hints to progressively guide toward the correct answers. MPR is orthogonal to CoT and self-consistency, making it easy to integrate with state-of-the-art techniques to further enhance performance\cite{zheng2023progressivehint}. Self Consistent Chain of Thought (CoT-SC)\cite{wang2023selfconsistency} is a technique that enhances the accuracy of the responses from large language models compared to the pure Chain of Thought. It operates under the assumption that the correct answer resides within the language model and that this correct answer is returned from the language model in the majority of instances while repeatedly posing the same question to the model. Tree of Thoughts (ToT)\cite{yao2023tree}is a paradigm that permits LLMs to navigate multiple reasoning paths when resolving problems. The architecture of ToT comprises four parts: Thought decomposition (segmenting a complex problem into smaller parts), thought generation (formulating potential solutions for each part), and state evaluation (assessing the potential solutions).
Graph of Thoughts (GoT)\cite{besta2023graph}, a framework that propels the prompting capabilities in LLMs beyond those offered by paradigms such as Chain-of-Thought or Tree of Thoughts (ToT). The fundamental idea and primary advantage of GoT is its ability to model the information generated by an LLM as an arbitrary graph, where units of information (“LLM thoughts”) serve as vertices, and edges correspond to dependencies between these vertices.

Environmental feedback can significantly enhance the efficacy of Large Language Model agents. “ReAct” (Reason for Future, Act for Now)\cite{yao2023react} is a methodological framework for autonomous Large Language Model (LLM) agents, demonstrating provable sample efficiency. Within this framework, an LLM agent contemplates the long-term implications of actions from the current state. Following a state transition in the external environment, the LLM agent re-engages the reasoning routine to strategize a new future trajectory from the updated state. This cyclical process facilitates the agent’s self-improvement through environmental feedback. Reflexion\cite{shinn2023reflexion} is another framework in which an LLM agent assimilates environmental feedback to refine its internal world model, thereby enhancing its ability to foresee the outcomes of its actions and make more judicious decisions.

For aligning the response of LLM agents with application preference, Human feedback is instrumental in the training of Large Language Models. It imparts vital information that may not be directly gleaned from environmental rewards. While rewards steer the agent towards the desired outcome, human feedback introduces a layer of expertise, aiding the agent in navigating intricate scenarios. For instance, in a chess game, a reward-based RL agent might learn to win games, but with human feedback, it can acquire strategies and tactics that would have otherwise required countless iterations to discover.

RLHF\cite{ziegler2019fine}is a machine learning methodology where a “reward model” is trained using direct human feedback, which is then employed to optimize the performance of an artificial intelligence agent via reinforcement learning. RLHF is particularly well-suited for tasks with goals that are complex, ill-defined, or challenging to specify. In contrast, Direct Preference Optimization (DPO)\cite{rafailov2023direct} offers an alternative to RLHF. DPO eliminates the need for fitting a reward model, sampling from the language model during fine-tuning, or conducting extensive hyperparameter tuning. DPO leverages a relationship between reward functions and optimal policies to address the reward maximization problem with constraints in a single policy training phase. Policy training directly resolves a classification problem on LLM-generated text using human feedback. Human feedback must be expressed as binary preferences between two alternatives. The classification problem involves predicting which alternative the user prefers, given the text generated by the LLM. DPO obviates the need to train a reward model, the sample from the LLM during fine-tuning or conduct an extensive hyperparameter search. Figure 2 illustrates the pivotal procedures involved in RLHF and DPO.

\begin{figure}[htp]
    \centering
    \includegraphics[scale=0.25]{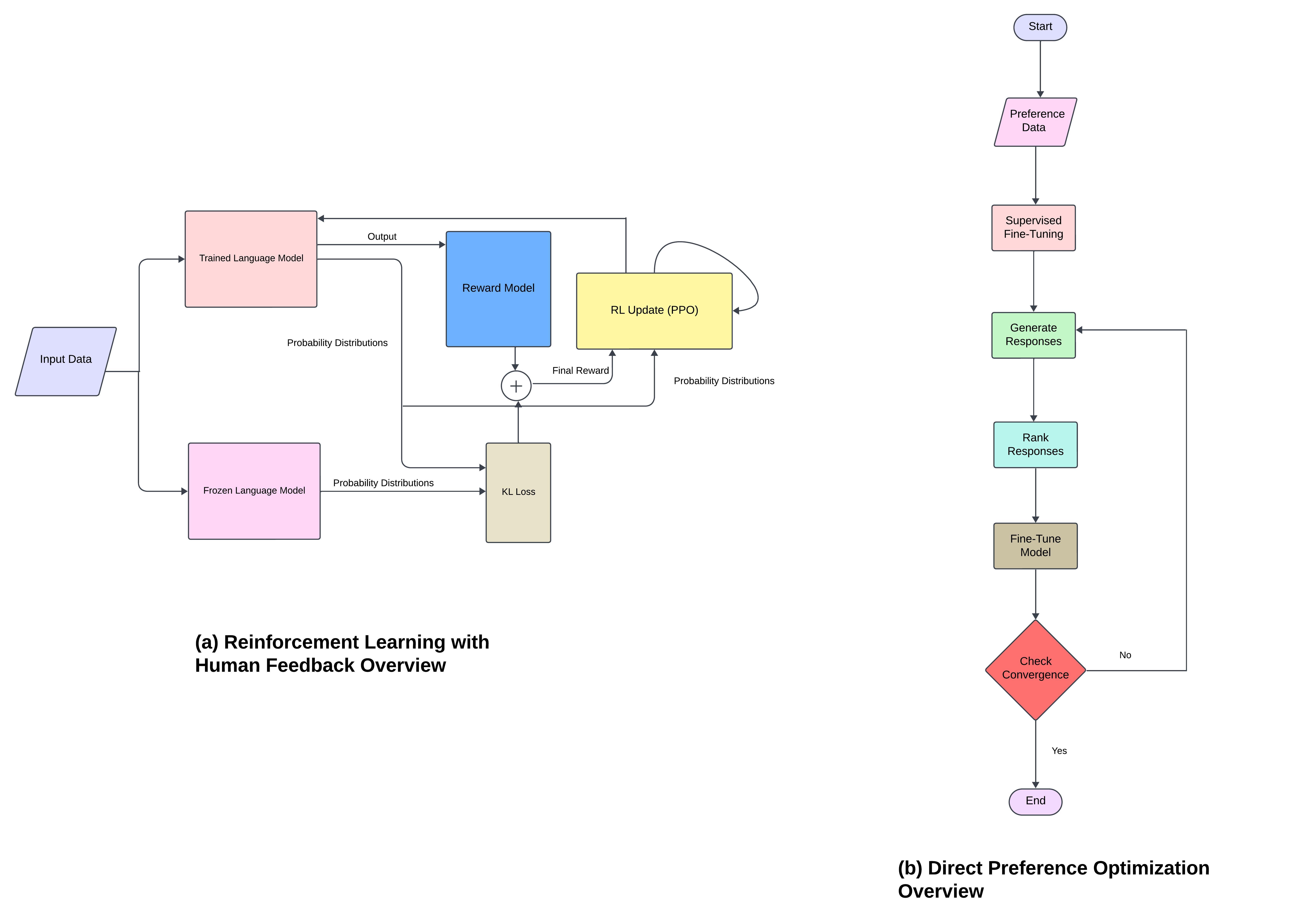}
    \caption{The Procedures of RLHF and DPO}
    \label{fig:my_label}
\end{figure}

Retrieval Augmented Generation (RAG) has emerged as a favored paradigm for enabling Large Language Models (LLMs) to access external data, serving as a grounding mechanism to counter hallucinations. RAG models amalgamate pre-trained parametric and non-parametric memory for language generation. The parametric memory is a pre-trained seq2seq model, while the non-parametric memory is a dense vector index of Wikipedia accessed with a pre-trained neural retriever. With RAG, LLMs retrieve contextual documents from a database to enhance the accuracy of their responses. Frameworks such as LangChain, LlamaIndex, FastRAG, and others serve as orchestrators, connecting LLMs with tools, databases, memories, etc., thereby augmenting their capabilities. User instructions are not inherently optimized for retrieval. Various techniques, including multi-query retriever, HyDE, etc., can be employed to rephrase/expand them and enhance performance. Figure 3 elucidates the operational mechanism of the RAG.

\begin{figure}[htp]
    \centering
    \includegraphics[scale=0.25]{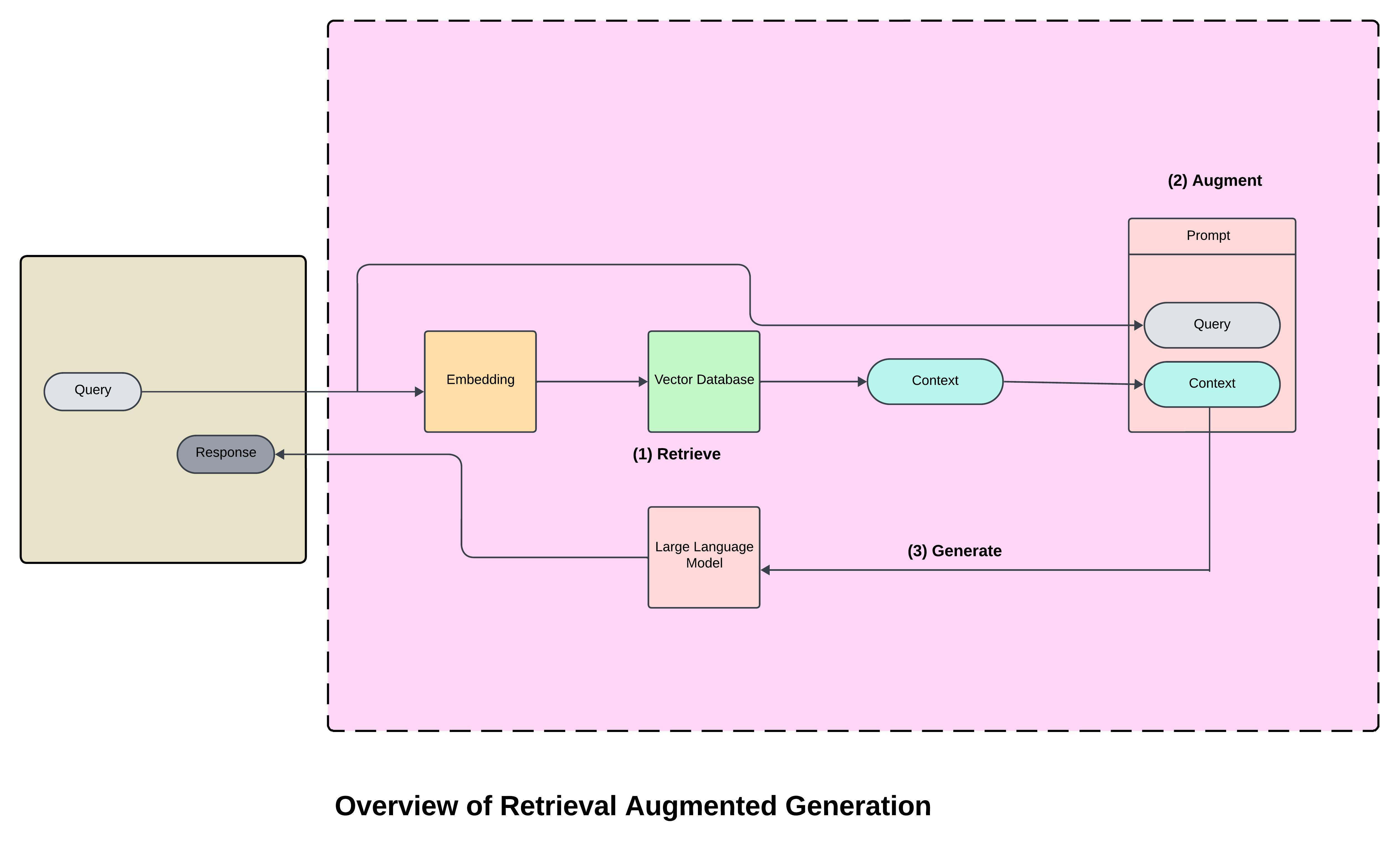}
    \caption{Overview of Retrieval Augmented Generation}
    \label{fig:my_label}
\end{figure}

To recall previous instructions and responses, LLMs and chatbots like ChatGPT incorporate this history into their context window. This buffer can be enhanced with summarization (e.g., using a smaller LLM), a vector store + RAG, etc\cite{zhong2023memorybank}. Both the document retrieval (context precision and recall) and generation stages (faithfulness and answer relevancy) need to be evaluated. Tools like Ragas and DeepEval can simplify this process\cite{gienapp2023evaluating}.

Advanced RAG techniques bolster retrieval techniques for RAG models and evaluate their performance using industry-standard metrics. Advanced RAG techniques systematize various approaches and provide a comprehensive examination of the progression of RAG paradigms. They encompass the Naive RAG, the Advanced RAG, and the Modular RAG. Pinecone's LLM Agent exemplifies an agent that can utilize tools like calculators, search, or executing code. Using agents, an LLM can write and execute Python code. It can search for information and even query a SQL database. LangChain offers a method to construct queries for LLMs. It employs a query-constructing LLM chain to write a structured query and then applies that structured query to its underlying VectorStore. LangChain enables LLMs to connect to a SQL database. Since LangChain uses SQLAlchemy to connect to SQL databases, we can use any SQL dialect supported by SQLAlchemy, such as MS SQL, MySQL, MariaDB, PostgreSQL, Oracle SQL, Databricks, or SQLite.

\subsection{Prompting is all you need}

With Refined Prompts, LLM Agent's Performance can be enhanced by
acquiring linguistic competencies in an unsupervised manner, predicting masked words within sentences. The efficacy of these LLMs hinges on the prompts or inputs they receive. The art of prompt engineering involves the creation of high-quality prompts or queries that are bespoke to the task at hand. By furnishing the model with lucid and pertinent prompts, it can more readily generate precise and germane responses. This process of refining and modifying the prompt to elicit a “superior” or more desired output from the model is a rudimentary illustration of prompt engineering.

Multiple Techniques to Augment the Prompting for a Model include: 

\begin{itemize}

    \item Automatic Reasoning and Tool Use (ART)\cite{paranjape2023art}: ART attains a substantial enhancement over few-shot prompting and automatic CoT on unseen tasks in the BigBench and MMLU benchmarks.
    
    \item Automatic Prompt Engineer (APE): APE autonomously generates instructions for a task that is delineated via output demonstrations. It formulates several instruction candidates, executes them using the target model, and selects the most suitable instruction based on computed evaluation scores.
    
    \item Active-Prompt\cite{diao2023active}: Active-Prompt is a technique that involves designing prompts that actively steer the model toward generating the desired output.
    
    \item Directional Stimulus Prompting\cite{li2023guiding}: This technique introduces a novel component, termed directional stimulus, into the prompt, providing more fine-grained guidance and control over LLMs.
    
    \item Program-Aided Language Models (PAL): PALs are capable of writing code that solves a question. They dispatch the code to a programmatic runtime to obtain the result.
    
    \item Multimodal CoT: Multimodal CoT integrates text and vision into a two-stage framework. The initial step involves rationale generation based on multimodal information. This is succeeded by the second phase, answer inference, which leverages the informative generated rationales.
    
    \item Graph Prompting\cite{wu2023survey}: Graph prompting involves using graph data, as graphs serve as structured knowledge repositories by explicitly modeling the interaction between entities.

\end{itemize}

Prompt tuning, a parameter-efficient tuning (PETuning) method, is employed for harnessing pre-trained models (PTMs). This technique simply appends a soft prompt to the input and solely optimizes the prompt to adapt PTMs to downstream tasks\cite{liu2022late}. The significance of this technique lies in its efficiency, as training a large language model from scratch is prohibitively expensive in terms of computational resources and time. By capitalizing on the knowledge already encapsulated in the pre-trained model, one can achieve high performance on specific tasks with considerably less data and computation. In the context of prompt-tuning, optimal cues or front-end prompts are fed to the AI model to provide it with task-specific context. These prompts can be additional words introduced by a human, or AI-generated numbers incorporated into the model’s embedding layer. Analogous to crossword puzzle clues, both types of prompts guide the model toward a desired decision or prediction. Prompt tuning enables a company with limited data to tailor a massive model to a narrow task. It also obviates the need to update the model’s billions (or trillions) of weights or parameters. Redeploying an AI model without retraining it can reduce computing and energy use by at least 1,000 times, resulting in substantial cost savings.

Prompt and Prefix Token Tuning is a technique that involves the optimization of continuous prompts for generation\cite{li2021prefixtuning}. In prefix-tuning, the context (more concretely, prompt embeddings) must be reintroduced with every request in conversations since the prompts do not alter the model. Prefix-tuning is a lightweight alternative to fine-tuning for natural language generation tasks, which maintains language model parameters frozen, but optimizes a small continuous task-specific vector (referred to as the prefix). Prefix-tuning draws inspiration from prompting, allowing subsequent tokens to attend to this prefix as if they were “virtual tokens”. By learning only 0.1\% of the parameters, prefix-tuning achieves comparable performance in the full data setting, outperforms fine-tuning in low-data settings, and extrapolates better to examples with topics unseen during training.

Dataset-specific prefixes in prompting pertain to the application of distinct prompts that are customized to the dataset in question\cite{huang2023diversityaware}. This technique is indispensable in scenarios where datasets exhibit substantial diversity, and a generic prompt per dataset may not adequately manage the intricate distribution shift towards the original pretraining data distribution. To address this, a dataset Diversity-Aware prompting strategy is proposed, which is initialized by a Meta-prompt. The downstream dataset is clustered into small homogeneity subsets in a diversity-adaptive manner, with each subset having its own separately optimized prompt. This divide-and-conquer design mitigates the optimization difficulty and significantly enhances the prompting performance.

The provision of pseudocode and code snippets as examples can markedly enhance the responses generated by a model\cite{mishra2023prompting}. Pseudocode is an informal high-level description of the operating principle of a computer program or an algorithm. It serves as a tool that programmers utilize to plan and organize their code. When a model is provided with pseudocode, it can comprehend the logic underpinning the code and generate a more accurate and pertinent response. This is because pseudocode offers a clear and concise representation of the algorithm, facilitating the model’s understanding and generation of the desired output.

\begin{figure}[htp]
    \centering
    \includegraphics[scale=0.5]{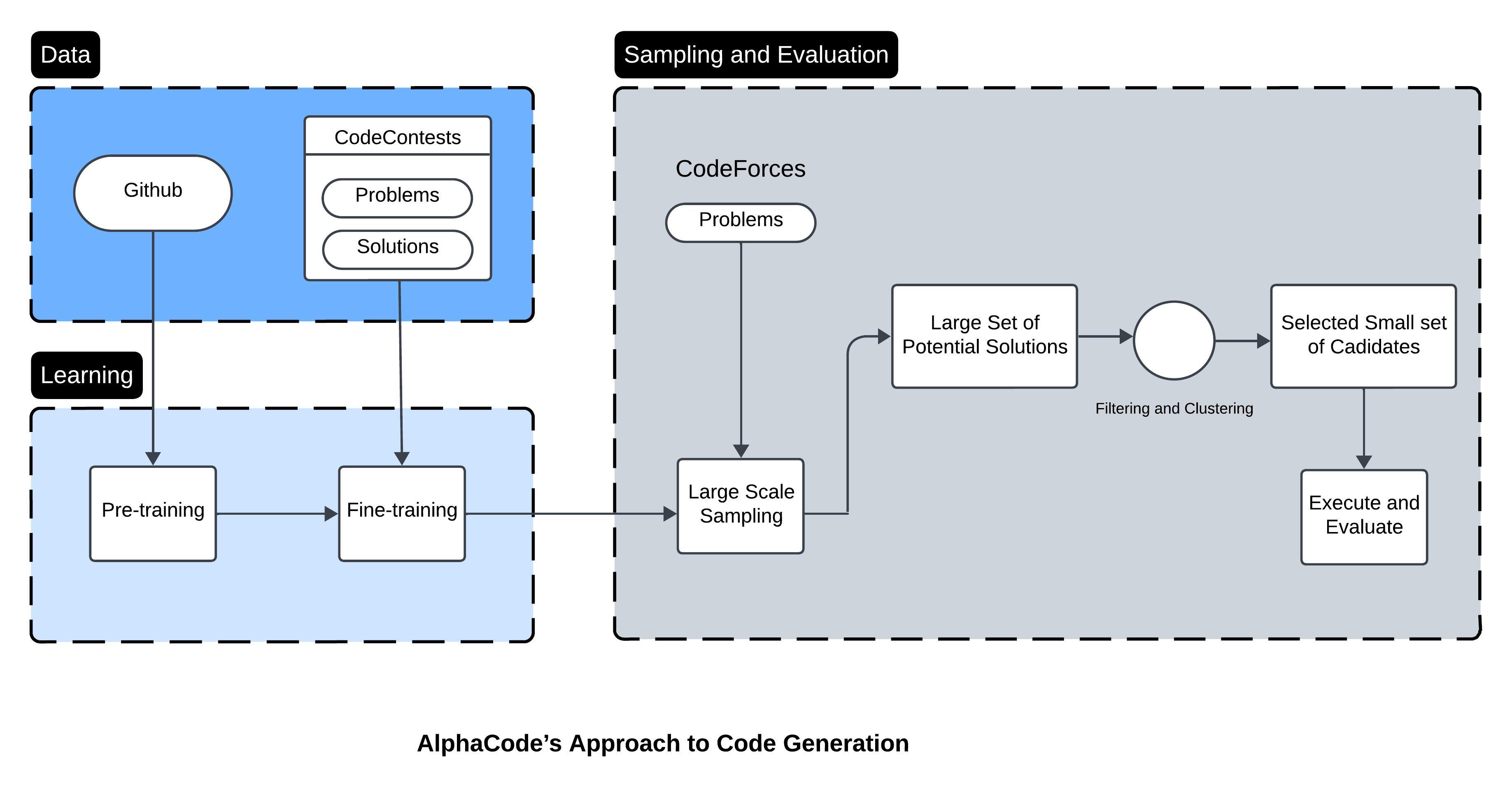}
    \caption{AlphaCode's Approach to Code Generation}
    \label{fig:my_label}
\end{figure}

In multimodal models, the provision of a Unified Modeling Language (UML) diagram as input can significantly enhance the generated code\cite{hu2024mplugpaperowl}. UML diagrams offer a standardized method to visualize a system’s architectural blueprints, encompassing elements such as activities, actors, business processes, database schemas, components, programming language statements, and reusable software components. When a multimodal model is provided with a UML diagram, it can comprehend the system’s architecture and generate code that accurately represents the system. koziolek et al.\cite{koziolek2023llmbased} explores such diagram-based code generation in LLMs. Figure 4 shows the methodology employed by AlphaCode\cite{Li_2022} for code generation.

In-context learning is a potent strategy for extracting knowledge from Large Language Models\cite{kirk2023improving}. It involves providing the model with a context or a situation, and the model learns to generate responses based on that context. This is particularly useful in scenarios where the model needs to adapt to new situations or tasks. For example, in a conversation, the model can use the context of the previous messages to generate a relevant response. In-context learning enables the model to continuously learn and adapt as it interacts with the environment, making it a powerful tool for tasks such as conversation, question answering, and more.

\begin{figure}[htp]
    \centering
    \includegraphics[scale=0.3]{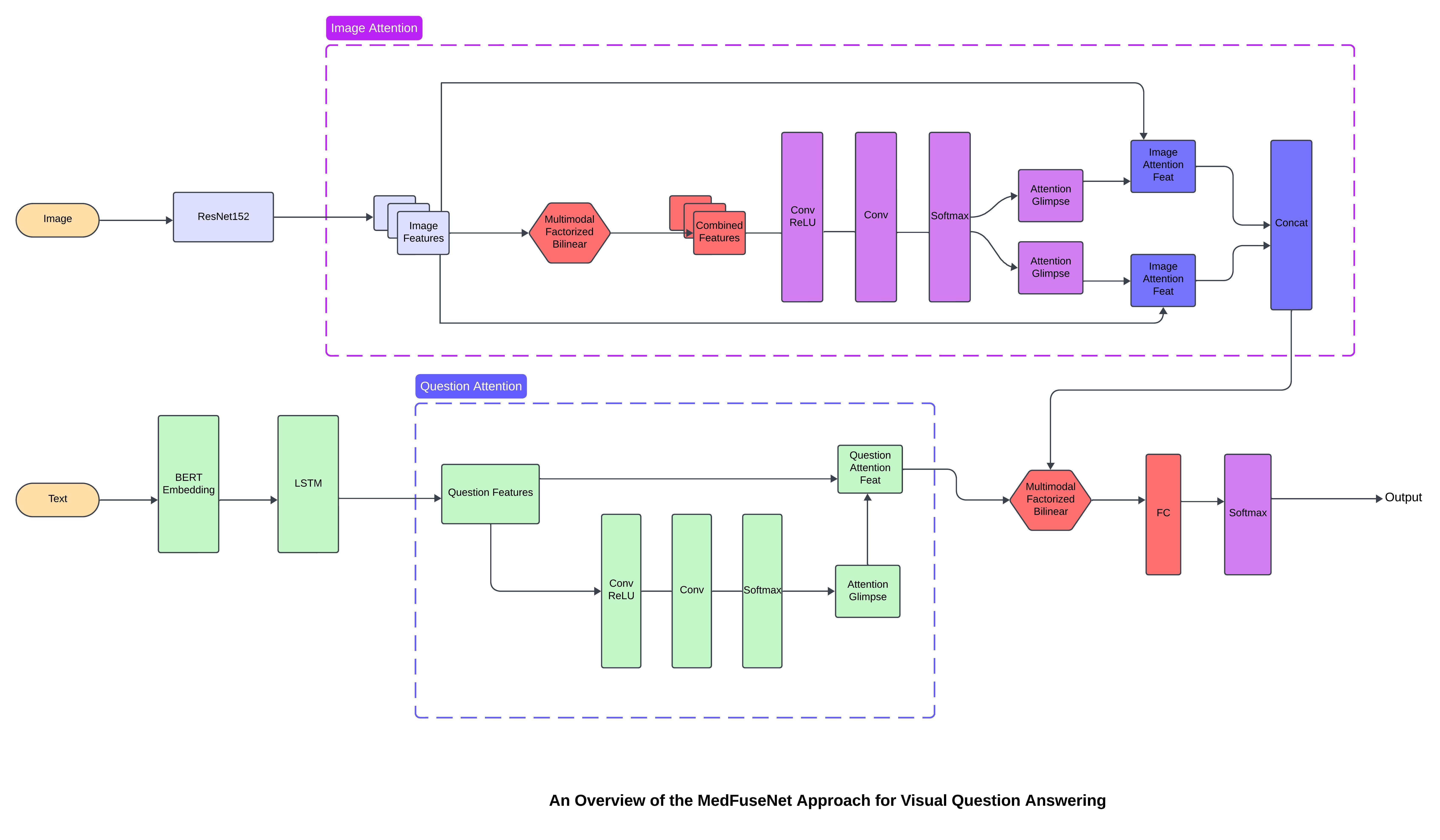}
    \caption{MedFuseNet's Approach to Visual Question Answering}
    \label{fig:my_label}
\end{figure}

Large Language Models (LLMs) can execute Visual Question Answering (VQA) by incorporating visual comprehension with language processing. The model is furnished with an image and a query about the image, and it formulates an answer grounded in its understanding of the image content and the query. A technique termed Img2Prompt\cite{guo2023from} is employed to offer prompts that delineate image content and self-constructed question-answer pairs, which effectively steer the LLM to execute VQA tasks. This technique is agnostic to LLMs and can operate with any LLM to execute VQA. An alternative approach is the Retrieval-Augmented Generation (RAG) for Visual Prompt Queries, where an intelligent search tool seeks pertinent information in the provided Knowledge bases. The retrieved information is subsequently fed to the LLM to augment its capacity to generate precise and relevant answers. Figure 5 visualizes MedFuseNet's\cite{sharma2021medfusenet} Approach to Visual Question Answering.

In protracted conversations, the application of prompts is indispensable to maintain the context and coherence of the dialogue. For dialogue applications that necessitate very long conversations, one strategy is to summarise or filter previous dialogue since models have a fixed context length. Researchers from Meta AI, MIT, and CMU have proposed an intriguing solution termed StreamingLLM\cite{xiao2023efficient}, which permits existing LLMs to handle extremely long contexts without any fine-tuning. An alternative approach is the ConversationChain, which enables a Large Language Model to recall previous interactions with the user. By default, LLMs are stateless — signifying each incoming query is processed independently of other interactions. The memory permits an LLM to recall previous interactions.

Prompting is also utilized in common sense reasoning tasks, where the model is anticipated to make inferences or judgments that are evident to humans but may not be explicitly stated\cite{wang2023recursively}. For instance, if someone uploads an image of a piece of furniture and provides instructions, the model can utilize these prompts to generate an advertisement for the furniture. Liu et al.\cite{liu2022generated} explores the application of generated knowledge prompting, which involves generating knowledge from a language model and then providing the knowledge as additional input when answering a question. This method enhances the performance of large-scale models on various commonsense reasoning tasks.

Prompting is a crucial aspect of utilizing Large Language Models effectively. It involves crafting clear and specific instructions or queries to elicit the desired responses from the language model\cite{friedman2023leveraging}. The LLM Retrieval Augmented Generation augments the content produced by LLMs through the addition of relevant material retrieved from external sources. The core strength of RAG lies in its method of data retrieval, providing LLMs with extra context that significantly enriches the content generation process.

Constructive prompt data generation is a method where prompts are meticulously crafted to steer AI models toward generating synthetic data. This artificially created data serves as a valuable resource for training and testing AI models, especially when real-world data for a specific task or domain is scarce. The spectrum of data generation techniques spans from simple methods such as data augmentation, which involves minor alterations to existing data, to more intricate methods like employing generative models to fabricate entirely new data samples. Ensuring Diversity in generated datasets is paramount for the training of robust and generalizable AI models\cite{wei2023jailbreak}. Techniques to enhance diversity encompass:

\begin{itemize}
    \item Utilizing various data generation methods like bootstrapping lays the groundwork for generating diverse models via resampling.

    \item Incorporating an assortment of data sources could entail collecting data from disparate databases or employing different types of data (e.g., text, images, audio) for a multi-modal approach.

    \item Data augmentation can be applied to amplify the size and diversity of an existing training set without manually collecting new data\cite{paper2021increase}. This process fabricates additional training data from existing examples by augmenting them using random transformations that yield plausible images. Regular evaluation of the dataset’s diversity and necessary adjustments are also crucial.
    
\end{itemize}

The prompt function in AI models plays a pivotal role in processing a user’s input and generating an appropriate prompt for the model. This involves deciphering the user’s input, identifying the desired task, and formulating a prompt that effectively conveys this task to the model. The objective is to steer the model’s output towards a specific function or behavior. This could involve specifying the format of the desired output, providing examples, instructing the model to think step-by-step, or debating the pros and cons before settling on an answer.

The responsible use of AI models necessitates ensuring that the models are used in a manner that is ethical, fair, and respectful of users’ privacy. This includes thwarting prompt injections, which involve manipulating the prompt to make the model behave in undesirable ways and rejecting harmful prompts, which could lead to the model generating inappropriate or harmful content\cite{friedman2023vendi}. It’s imperative to have safeguards in place to detect and prevent such misuse and to regularly review and update these safeguards as necessary. For instance, one approach to prevent prompt injections is to enhance the robustness of the internal prompt that is added to the user input. Other potential safety measures, as per Perez et al.\cite{perez2022ignore}, include setting a lower temperature and increasing the frequency penalty. Furthermore, since elaborate prompt injections may require a substantial amount of text to provide context, simply limiting the user input to a reasonable maximum length makes prompt injection attacks significantly more challenging.

\section{Evaluating Autonomous Agents}

The evaluation of autonomous agents is a pivotal component in their development and deployment process. It ensures that these agents perform as anticipated under a myriad of foreseen and unforeseen conditions. The evaluation process is indispensable in ensuring that engineered systems, such as autonomous agents, meet the desired performance benchmarks and navigate the complexities of real-world scenarios. At their essence, autonomous agents are intelligent entities capable of making decisions and executing actions without direct human intervention. They utilize advanced algorithms and machine learning models to analyze data, derive insights, and carry out tasks autonomously. Hence, evaluating these agents is vital to ensure they make appropriate decisions and execute tasks effectively.
Furthermore, evaluation is crucial for identifying and rectifying potential issues or limitations in the agent’s performance. It enables developers to monitor the agent’s performance, pinpoint areas for improvement, and implement necessary adjustments to enhance its effectiveness. Additionally, evaluation plays a significant role in fostering trust in autonomous agents. By demonstrating that these agents can reliably and accurately perform tasks, evaluation aids in building user trust and acceptance. This is particularly crucial for autonomous agents deployed in high-stakes contexts, where their decisions and actions can have substantial consequences.

\subsection{Traditional Evaluation Frameworks}

Traditional evaluation frameworks for autonomous agents typically focus on assessing the agent’s performance in isolated environments. These frameworks generally involve testing the agent’s capability to complete specific tasks or achieve certain objectives within a controlled environment. A common approach is to use benchmark tasks or datasets to evaluate the agent’s performance\cite{kinniment2024evaluating}. These benchmarks provide a standard measure of performance that can be used to compare different agents or algorithms. Another prevalent approach is to use simulation environments to evaluate the agent’s performance. These environments enable developers to test the agent’s performance under various conditions and scenarios\cite{9497472}.

However, these traditional frameworks often overlook the unique challenges of evaluating autonomous agents. For instance, they may not adequately consider the dynamic environments in which these agents operate or the complex interactions between agents and their environment. Therefore, new evaluation frameworks are being developed to address these challenges. These frameworks aim to incorporate evaluation throughout the development life cycle and into operation as the system learns and adapts in a noisy, changing, and contended environment. They account for the challenges of testing the integration of diverse systems at various hierarchical composition scales while respecting that testing time and resources are limited\cite{Lanus_2021}.

\subsection{AgentBench: An All-Encompassing Evaluation Framework}

AgentBench\cite{liu2023agentbench} is an evolving, multi-dimensional benchmark currently comprising eight unique environments. It is designed to evaluate the reasoning and decision-making capabilities of Large Language Models in a multi-turn, open-ended generation setting. These environments cover a broad spectrum of domains, each posing distinct challenges and requirements for the LLMs. Five of these environments are newly created, specifically, the operating system, database, knowledge graph, digital card game, and lateral thinking puzzles. Each environment represents a different facet of real-world tasks that an LLM might encounter. For example, the OS environment assesses the LLM’s ability to interact with a simulated operating system, while the digital card game environment evaluates the LLM’s strategic decision-making abilities in a card game. In addition to these novel environments, AgentBench also incorporates three environments recompiled from published datasets: House-Holding (HH), Web Shopping (WS), and Web Browsing (WB). These environments offer a diverse array of challenges, ranging from managing household tasks in the HH environment to navigating online shopping platforms in the WS environment.

\subsubsection{LLMs Performance in AgentBench}

The authors carried out extensive tests over 27 API-based and open-sourced (OSS) LLMs. The results revealed a significant performance gap between top commercial LLMs and their OSS counterparts. While the top commercial LLMs exhibited a robust ability to act as agents in complex environments, the OSS LLMs fell short. The authors identified several common reasons for failures in both the environments and the LLMs. They discovered that poor long-term reasoning, decision-making, and instruction-following abilities were the primary hurdles for developing usable LLM agents. These findings underscore the challenges in developing LLMs that can effectively navigate complex environments and perform tasks over multiple turns.

\subsubsection{Enhancing Agent Performance}

The authors propose that training on code and high-quality multi-turn alignment data could enhance agent performance. By training on code, LLMs can gain a better understanding of the structure and logic of programming languages, which is particularly beneficial for environments like OS and DB. On the other hand, high-quality multi-turn alignment data can assist LLMs in better understanding the context of a conversation and making more appropriate decisions. In addition to these proposals, the authors also released datasets, environments, and an integrated evaluation package for AgentBench. These resources can be utilized by other researchers to evaluate their own LLMs and contribute to the ongoing development of LLMs as agents.

\subsubsection{Comparison with Traditional Evaluation Frameworks}

This approach differs from traditional evaluation frameworks in several ways. Traditional evaluation frameworks often focus on assessing the agent’s performance in isolated environments. These frameworks typically involve testing the agent’s ability to complete specific tasks or achieve certain objectives within a controlled environment. However, these traditional frameworks often fail to account for the unique challenges of evaluating autonomous agents. For instance, they may not adequately account for the dynamic environments in which these agents operate, or the complex interactions between agents and their environment. AgentBench, on the other hand, provides a more comprehensive evaluation of the LLMs’ ability to operate as autonomous agents in various scenarios. It encompasses a diverse spectrum of different environments, providing a more realistic and challenging setting for evaluating the agents. This makes it a more effective tool for assessing the performance of LLMs as agents and identifying areas for improvement.

\subsection{WebArena: A Novel Environment for Autonomous Agents}

WebArena\cite{zhou2023webarena} is a highly realistic and reproducible environment designed for language-guided agents. It aims to bridge the gap between the simplified synthetic environments where current agents are primarily developed and tested, and the complex real-world scenarios they are expected to navigate. The environment encompasses fully functional websites from four common domains: e-commerce, social forum discussions, collaborative software development, and content management. These domains represent a broad spectrum of tasks that humans regularly perform on the internet, thus providing a diverse and challenging setting for evaluating agents. In addition to the websites, the environment is enriched with tools (e.g., a map) and external knowledge bases (e.g., user manuals) to promote human-like task-solving. This feature is designed to push the boundaries of what LLMs can achieve, moving beyond simple question-answering or text-generation tasks to more complex problem-solving tasks.

\subsubsection{Benchmark Tasks and Evaluation}

Building on the WebArena environment, the authors release a set of benchmark tasks focusing on evaluating the functional correctness of task completions. These tasks are diverse, long-horizon, and designed to emulate tasks that humans routinely perform on the internet. The authors conducted extensive tests over several baseline agents, integrating recent techniques such as reasoning before acting. The results demonstrate that solving complex tasks is challenging: their best GPT-4-based agent only achieves an end-to-end task success rate of 14.41\%, significantly lower than the human performance of 78.24\%. These results underscore the need for further development of robust agents and indicate that current state-of-the-art large language models are far from perfect performance in these real-life tasks. They also demonstrate that WebArena can be used to measure such progress.

\subsubsection{Comparison with Traditional Approaches}

The approach taken by WebArena significantly diverges from traditional approaches to agent development and evaluation. Traditional approaches often involve creating and testing agents in simplified synthetic environments. While these environments are useful for initial development and testing, they often fail to accurately represent the complexity and diversity of real-world scenarios. In contrast, WebArena provides a highly realistic and reproducible environment for developing and testing agents. By including fully functional websites from common domains and enriching the environment with tools and external knowledge bases, WebArena provides a setting that closely mirrors the real-world tasks that humans routinely perform on the internet.

This approach allows for a more comprehensive evaluation of the agents’ performance and capabilities. It also provides a platform for identifying and addressing the challenges and limitations of current LLMs, paving the way for the development of more robust and capable agents.

\subsection{ToolLLM: Facilitating Large Language Models to Master 16000+ Real-world APIs}

ToolLLM\cite{qin2023toolllm} introduces a comprehensive tool-use framework that includes data construction, model training, and evaluation. This framework is designed to enhance the capabilities of Large Language Models (LLMs) in using external tools (APIs) to fulfill human instructions. This is a crucial aspect of LLMs’ functionality, as it allows them to interact with the real world and perform complex tasks.

Despite the advancements of open-source LLMs, such as LLaMA, their tool-use capabilities remain significantly limited. This is primarily because current instruction tuning largely focuses on basic language tasks and overlooks the tool-use domain. This stands in stark contrast to the excellent tool-use capabilities of state-of-the-art (SOTA) closed-source LLMs, such as ChatGPT.

\subsubsection{Data Construction, Model Training, and Evaluation}

To address this gap, the authors of ToolLLM introduce ToolBench, an instruction-tuning dataset for tool use, which is automatically constructed using ChatGPT. The construction process can be divided into three stages:

\begin{itemize}
    \item API collection: The authors collect 16,464 real-world RESTful APIs spanning 49 categories from RapidAPI Hub.
    \item Instruction generation: They prompt ChatGPT to generate diverse instructions involving these APIs, covering both single-tool and multi-tool scenarios.
    \item Solution path annotation: They use ChatGPT to search for a valid solution path (chain of API calls) for each instruction.
\end{itemize}

To enhance the reasoning capabilities of LLMs, the authors develop a novel depth-first search-based decision tree algorithm. It enables LLMs to evaluate multiple reasoning traces and expand the search space. Moreover, to evaluate the tool-use capabilities of LLMs, the authors develop an automatic evaluator: ToolEval. Based on ToolBench, they fine-tune LLaMA to obtain an LLM ToolLLaMA, and equip it with a neural API retriever to recommend appropriate APIs for each instruction.

\subsubsection{Performance of ToolLLaMA}

Experiments show that ToolLLaMA demonstrates a remarkable ability to execute complex instructions and generalize to unseen APIs, and exhibits comparable performance to ChatGPT. ToolLLaMA also demonstrates strong zero-shot generalization ability in an out-of-distribution tool-use dataset: APIBench.

\subsubsection{Comparison with Traditional Approaches}

The approach taken by ToolLLM significantly diverges from traditional approaches to agent development and evaluation. Traditional approaches often involve creating and testing agents in simplified synthetic environments. While these environments are useful for initial development and testing, they often fail to accurately represent the complexity and diversity of real-world scenarios.In contrast, ToolLLM provides a highly realistic and reproducible environment for developing and testing agents. By including fully functional websites from common domains and enriching the environment with tools and external knowledge bases, ToolLLM provides a setting that closely mirrors the real-world tasks that humans routinely perform on the internet. This approach allows for a more comprehensive evaluation of the agents’ performance and capabilities. It also provides a platform for identifying and addressing the challenges and limitations of current LLMs, paving the way for the development of more robust and capable agents.

\subsection{Subjective Evaluation in LLM-Based Autonomous Agents}

Subjective evaluation plays a pivotal role in the development and deployment of Large Language Models (LLMs) based autonomous agents. This process involves gauging the performance of these agents through human judgment, offering valuable insights into their effectiveness, usability, and overall quality.

One prevalent method employed in the subjective evaluation of LLMs is human annotation\cite{schwartz2023enhancing}\cite{wang2023survey}. This process entails human annotators reviewing and rating the outputs of the LLMs based on various criteria such as relevance, coherence, and fluency. This method can yield a detailed assessment of the LLM’s performance and pinpoint areas that may require improvement. Nonetheless, Human annotation comes with its own set of challenges. It can be time-consuming and expensive, particularly for large-scale evaluations. Furthermore, it can be subject to bias, as different annotators may interpret the evaluation criteria differently. Despite these challenges, human annotation remains an invaluable tool for the subjective evaluation of LLMs.

The Turing Test, named after the British mathematician and computer scientist Alan Turing, is another method used in the subjective evaluation of LLMs. The Turing Test is designed to assess a machine’s ability to exhibit intelligent behavior equivalent to, or indistinguishable from, that of a human. In the context of LLMs, the Turing Test typically involves a human judge engaging in a conversation with the LLM without the knowledge that they are interacting with a machine\cite{feldt2023autonomous}\cite{wang2023survey}. If the judge cannot reliably distinguish the LLM from a human interlocutor, the LLM is deemed to have passed the Turing Test. However, the Turing Test has its limitations. While it provides a high-level assessment of an LLM’s ability to mimic human-like conversation, it does not offer detailed insights into the specific strengths and weaknesses of the LLM. Moreover, passing the Turing Test does not necessarily imply that the LLM comprehends the content of the conversation in the same manner as a human would.

Beyond human annotation and the Turing Test, there exist other methods for the subjective evaluation of LLMs. For instance, user studies can be conducted to assess the usability and user satisfaction of LLMs in real-world scenarios\cite{xi2023rise}. These studies can offer valuable feedback on the LLM’s performance and identify areas for improvement. Another method involves the use of expert reviews, where experts in the field evaluate the LLM’s performance based on their professional judgment\cite{huang2024understanding}. This method can provide a more nuanced assessment of the LLM’s capabilities, especially in specialized domains.

\section{Constraints of Implementation }

\subsection{Multimodality: A Double-Edged Sword for LLM-Based Autonomous Agents}

The integration of diverse data types such as text, images, and audio, poses a formidable challenge\cite{koh2024visualwebarena}. The sophisticated processing required for multimodal data can strain the performance of these agents. The computational burden of multimodal data processing is a primary concern. Each data type necessitates distinct preprocessing steps, and the amalgamation of these disparate data types is computationally demanding. Moreover, the heterogeneity of data types engenders inconsistencies in the agent’s comprehension. For example, an agent may interpret textual data in a manner that diverges from its interpretation of visual data, potentially leading to decision-making conflicts. Additionally, the dearth of comprehensive multimodal datasets for training can curtail the agent’s capacity to understand and interpret multimodal data effectively. This limitation can render agents less adept at managing real-world scenarios, which often involve intricate, multimodal inputs.

As the call for multimodal capabilities in LLM-based autonomous agents intensifies, so does the pressure to deliver high-performing, reliable agents\cite{guo2024large}. However, this escalating demand can precipitate underperformance in several ways. The haste to incorporate multimodal capabilities can result in ill-conceived or poorly executed features, leading to suboptimal performance. Furthermore, the growing complexity of multimodal data can inundate the agent’s processing capabilities, resulting in slower response times and diminished efficiency. The broadening range of tasks that multimodal agents are expected to undertake can surpass their current capabilities, leading to errors and failures.

Several strategies can be employed to navigate the challenges associated with multimodality in LLM-based autonomous agents\cite{fan2024embodied}. Enhancing computational efficiency is paramount. This can be achieved through the use of optimized algorithms and more efficient data processing techniques. Besides, the development of more comprehensive and diverse multimodal datasets for training can bolster the agent’s ability to understand and interpret multimodal data. Meticulous design and implementation of multimodal features can prevent the introduction of poorly performing features\cite{fu2024limsim}.
A gradual expansion of the agent’s task scope, coupled with continuous testing and refinement, ensures the agent remains capable and reliable as it takes on more complex tasks.

\subsection{Human Alignment in LLM-Based Autonomous Agents}

The alignment of Large Language Model (LLM) based autonomous agents with human values, expectations, and instructions is a critical aspect of their design and operation\cite{wang2023aligning}. However, achieving this alignment is fraught with challenges that can adversely affect the performance of these agents\cite{tian2024evil}.
A common issue is the misunderstanding of human instructions. Subtle variations in the input prompt can lead to significant deviations in the output, resulting in actions that diverge from the user’s intentions. Additionally, LLM-based agents can inadvertently generate content that reflects biases present in the vast amounts of web data they are trained on. If not properly mitigated during training, these biases can manifest in the agent’s behavior. Furthermore, LLM-based agents can produce information that is either factually incorrect or semantically nonsensical. This can occur when the model makes erroneous inferences from the input data or generates grammatically correct but semantically meaningless text.

Misalignment between LLM-based autonomous agents and human expectations transpire in several ways. If an agent misinterprets a user’s instructions, it may take actions that are unhelpful or even counterproductive, leading to a decline in the agent’s overall effectiveness. The generation of biased content can erode the user’s trust in the agent. If users perceive the agent as biased, they may be less inclined to use it, or they may use it in a manner that reinforces their own biases. The production of incorrect or nonsensical information can sow confusion and misinformation, potentially leading users to make decisions based on incorrect information, with potentially serious consequences.

Various methods can be implemented to tackle the issues related to aligning with human values in LLM-based autonomous agents. Improving the quality of the training data can help mitigate misunderstandings and biases. This can involve the use of more diverse and representative datasets and the implementation of techniques to filter out biased or incorrect information. Incorporating user feedback usually aids in improving alignment. Users can provide valuable insights into the extent to which the agent’s actions align with their expectations, and this feedback can be used to further fine-tune the model\cite{arabzadeh2024better}. Ongoing monitoring and evaluation of the agent’s performance can help identify and rectify issues of misalignment. This can involve the use of metrics that specifically measure alignment and the implementation of systems for regular review and adjustment of the agent’s action.

\subsection{The Enigma of Hallucinations}

Hallucinations in Large Language Models (LLMs) are characterized by the model’s creation of content that lacks substantiation from its training data. This predicament can pose a significant obstacle to the performance of autonomous agents that utilize LLMs. The manifestation of hallucinations can lead to the generation of information that is either factually incorrect or devoid of logic. This can result in the agent delivering responses that are misleading or lack productivity, thereby undermining its overall efficiency. The presence of hallucinations can also introduce inconsistencies in the agent’s outputs. For instance, the agent may create content that contradicts information it previously generated or contradicts established facts. This can engender a state of confusion and foster distrust among users. Furthermore, hallucinations can prompt the agent to generate content that exhibits bias or is inappropriate. This can compromise the user’s trust in the agent and curtail its overall usefulness\cite{guan2023mitigating}.

Hallucinations can precipitate underperformance in LLM-based autonomous agents in several ways. The fabrication of incorrect or nonsensical information can lead to a decline in the overall effectiveness of the agent. Users may receive incorrect information, which can lead to misinformed decision-making. Hallucinations can damage the user’s trust in the agent\cite{yao2023llm}. If users perceive the agent as unreliable due to its hallucinations, they may be less inclined to use it, thereby reducing its utility. Hallucinations can lead to the agent fabricating inappropriate or offensive content. This can damage the user’s experience and may lead to reputational damage for the entity deploying the agent.

There are numerous approaches that can be utilized to tackle the difficulties related to hallucinations in autonomous agents based on LLM.
Enhancing the quality of the training data can help mitigate hallucinations. This can involve using more diverse and representative datasets, as well as implementing techniques to filter out biased or incorrect information. Incorporating feedback from users can help enhance the model’s performance. Users can provide valuable insights into the agent’s behavior, and this feedback can be used to fine-tune the model.
Ongoing monitoring and evaluation of the agent's performance can help identify and address issues of hallucinations\cite{ahmad2023creating}. This can involve using metrics that specifically measure the agent’s hallucination rate, as well as implementing systems for regular review and adjustment of the agent’s behavior. While hallucinations present significant challenges for LLM-based autonomous agents, these challenges can be surmounted with careful planning, efficient design, and continuous refinement. As the field continues to evolve, these agents are anticipated to become increasingly adept at handling complex tasks without hallucinating.

\subsection{The Intricacies of Agent Ecosystems}

The agent ecosystem, which refers to the environment in which autonomous agents based on Large Language Models (LLMs) operate, can significantly impact the performance of these agents. The intricacy of the agent ecosystem invokes computational inefficiencies. Each agent within the ecosystem necessitates distinct processing steps, and the interplay of these disparate agents can be computationally demanding\cite{nascimento2023selfadaptive}. Besides, the heterogeneity of agents within the ecosystem can induce inconsistencies in the system's overall performance. For instance, an agent might interpret data divergently from another agent, leading to potential conflicts in decision-making. The scarcity of comprehensive datasets for training can curtail the agent's capacity to comprehend and interpret the ecosystem effectively\cite{tang2024prioritizing}. This can result in less adept agents at managing real-world scenarios that often involve complex, multi-agent inputs.

The complexities of the agent ecosystem can lead to performance issues. If an agent cannot effectively navigate the ecosystem, it may undertake unhelpful or counterproductive actions. This can lead to a decline in the agent's overall effectiveness. Generating inconsistent responses typically erodes the user's trust in the agent\cite{lu2024agentlens}. If users perceive the agent as unreliable due to its inability to consistently interact with other agents, they may be less inclined to use it. The production of incorrect or nonsensical information sows confusion and misinformation\cite{tang2024prioritizing}. This can result in users making decisions based on incorrect information, with potentially serious consequences. In the meticulous design and standardization of LLM-based autonomous agents, standardized communication protocols must be implemented to ensure smooth information exchange and minimize misinterpretations. Each agent within the ecosystem should have clearly defined roles and responsibilities to prevent conflicts and redundancy. A modular design principle should be adopted, allowing for individual agents' independent development, testing, and deployment, thereby facilitating easier maintenance and scalability. For robust training and development, comprehensive training datasets that encompass diverse scenarios and data modalities (text, image, audio) should be developed to enhance the agents’ ability to handle real-world complexities. Mechanisms should be implemented for agents to continuously learn and adapt from their interactions with the environment and other agents, fostering their ability to handle unforeseen situations. Rigorous testing and refinement of the agent ecosystem should be conducted throughout the development process to identify and address potential issues before real-world deployment. Regarding trust and explainability, mechanisms that allow users to understand the rationale behind the agents’ decisions should be integrated, fostering trust and user acceptance. Techniques should be implemented to detect and mitigate potential biases within the agents and the training data to ensure fair and ethical behavior. Lastly, a framework should be maintained for human oversight and control over the agent ecosystem, allowing for intervention in critical situations and ensuring adherence to ethical guidelines.

\section{Conclusion}

Large Language Models (LLMs) are at the cutting edge of artificial intelligence, underpinning autonomous agents that are proficient in a broad spectrum of tasks across various domains. These agents, with their ability to comprehend and generate text akin to human communication, hold the potential to revolutionize sectors from customer service to healthcare.
Nonetheless, these agents grapple with several challenges. Multimodality, the capacity to process and generate information across diverse communication modes such as text, images, and sound, is a primary hurdle. While LLMs demonstrate excellence in text-based tasks, their efficacy in tasks involving other communication modes is yet to reach its peak. Another significant challenge is the alignment with human values. As AI systems gain more autonomy, it becomes imperative that their actions and decisions resonate with human ethics and values. This complex issue necessitates understanding cultural subtleties and ethical principles, and incorporating these into the AI system. The phenomenon of hallucinations, or the generation of ungrounded information, poses another obstacle. While LLMs are adept at producing plausible text, they occasionally generate information that is either factually incorrect or nonsensical. The evaluation of these agents also presents a considerable challenge. Conventional evaluation metrics such as accuracy or precision may not fully encapsulate the capabilities of these agents. Novel evaluation frameworks are required to assess their performance in intricate, real-world scenarios. To surmount these challenges, a variety of techniques are under exploration. Prompting and reasoning can steer the agent's responses 
and enhance their decision-making capabilities. The use of tools allows the agent to harness external resources to boost their capabilities. In-context learning empowers the agent to learn from the conversation history and tailor its responses accordingly. Robust evaluation platforms like AgentBench, WebArena, and ToolLLM offer comprehensive methods to assess these agents' performance in complex scenarios. These platforms emulate real-world environments and tasks, providing a thorough evaluation of the agent's capabilities. These advancements are charting the course for the development of more robust and capable autonomous agents. As research advances, we can anticipate these agents becoming an integral part of our digital lives, assisting us in tasks as diverse as responding to emails to diagnose diseases. The future of AI, with LLMs leading the charge, appears promising.

\newpage
\bibliographystyle{unsrt}
\bibliography{main.bib}

\end{document}